\documentclass[letterpaper]{article} 

\usepackage[dvipsnames]{xcolor}
\usepackage{amsmath}
\usepackage[most]{tcolorbox}
\usepackage{mathrsfs}
\usepackage[]{mdframed}
\usepackage{enumitem}
\usepackage{lipsum} 
\usepackage[utf8]{inputenc}
\usepackage{fourier} 
\usepackage{array}
\usepackage{makecell}
\usepackage{bm}
\usepackage{subcaption} 
\usepackage{lineno} 
\usepackage{scalerel}
\usepackage{tikz}
\usetikzlibrary{shapes.multipart} 
\usetikzlibrary{calc}
\usepackage{tabularx}
\usepackage{changepage,threeparttable} 
\usepackage{pifont}

\newcolumntype{b}{X}
\newcolumntype{s}{>{\hsize=.5\hsize}X}

\usepackage{booktabs, multirow} 
\usepackage{soul}

\newcommand{\fgreen}[1]{\textcolor{ForestGreen}{#1}}
\newcommand{\bred}[1]{\textcolor{BrickRed}{#1}}
\newcommand{\rblue}[1]{\textcolor{RoyalBlue}{#1}}

\usepackage{aaai25}  
\usepackage{times}  
\usepackage{helvet}  
\usepackage{courier}  
\usepackage[hyphens]{url}  
\usepackage{graphicx} 
\usepackage{natbib}  
\usepackage{caption} 
\usepackage{algorithm}
\usepackage{algorithmic}

\usepackage{newfloat}
\usepackage{listings}
\DeclareCaptionStyle{ruled}{labelfont=normalfont,labelsep=colon,strut=off} 
\floatstyle{ruled}
\newfloat{listing}{tb}{lst}{}
\floatname{listing}{Listing}
\title{Alexpaca: Learning Factual Clarification Question Generation Without Examples}
\author{
    Matthew Toles\textsuperscript{\rm 1},
    Yukun Huang\textsuperscript{\rm 2},
    Zhou Yu\textsuperscript{\rm 1},
    Luis Gravano\textsuperscript{\rm 1}
}
\affiliations{
    \textsuperscript{\rm 1}Department of Computer Science, Columbia University, New York, NY, USA\\
    \textsuperscript{\rm 2}Department of Computer Science, Duke University, Durham, NC, USA\\
    m.toles@columbia.edu, yukun.huang@duke.edu, yh3310@columbia.edu, gravano@cs.columbia.edu

}

\usepackage{bibentry}

\begin{document}

\maketitle


\begin{abstract}
    Real-life tasks such as giving legal or technical advice often lack complete context at the outset and can have disparate answers depending thereon. 
    The ability to derive missing factual information by asking clarifying questions (ACQ) is an important element of real-life collaboration on such reasoning tasks.
    Existing factual clarification question challenges evaluate generations based on word overlap or human evaluations.
    Recent work explores generating a response to the clarifying question then evaluating its utility directly.
    So far, these tasks are limited to disambiguating the user's intent rather than concrete facts about the situation.
    The factual domain presents unique challenges since responses to clarification questions must be factually true for accurate evaluation.
    To enable evaluation of factual domain clarification question generation, We present a new task that focuses on the ability to elicit missing information in multi-hop reasoning tasks.
    The task, HotpotQA-FLM, can be evaluated automatically, making it convenient for benchmarking language models.
    We observe that humans outperform GPT-4 by a large margin, while Llama 3 8B Instruct does not even beat the dummy baseline in some metrics.
    Finally, we find by fine-tuning Llama 3 8B Instruct on its own generations, filtered via rejection sampling, we can improve information recovery by 27.6\%. 
\end{abstract}

%

\section{Introduction}


In many real-world scenarios, the initial context is often incomplete, making it risky to provide answers without first seeking clarification. For instance, legal, medical, and technical advice typically depends on specific details about the individual's situation. As language models (LMs) are increasingly used in open-domain assistant roles, their ability to clarify and gather relevant facts before offering advice is becoming more crucial.

Evaluating clarification question generation is not straightforward.
Many question generation tasks evaluate generated questions based on word overlap with a ground truth label \cite{rahmani2023survey}, ignoring whether the question actually acquires useful information or how difficult it is to answer.
Word overlap also disregards that many different questions may be useful in a given situation, and a single question may be phrased in many different ways.
For example "Was Beethoven born before 1800" may be much easier for a person to recall than "What is Beethoven's birthday" even though there is little word overlap.


\begin{figure}[t]
    \centering
    \includegraphics[width=0.45\textwidth]{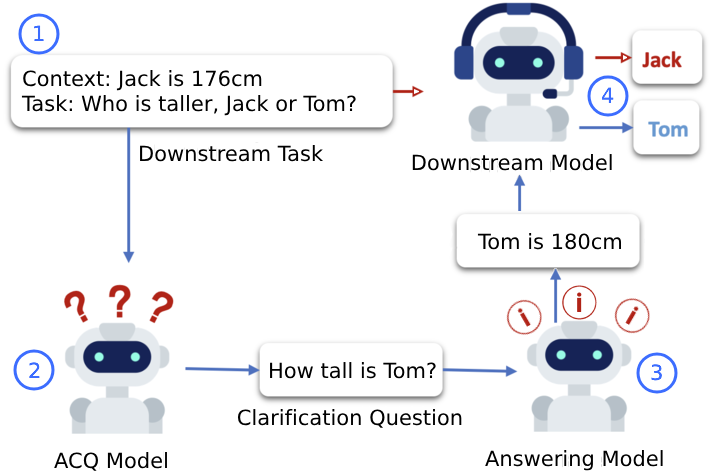}
    \caption{
        Overview of the HotpotQA-FLM task, which simulates the need to formulate a question. 
        Conventionally, the downstream model performs the downstream task directly (\protect\scalerel*{\includegraphics{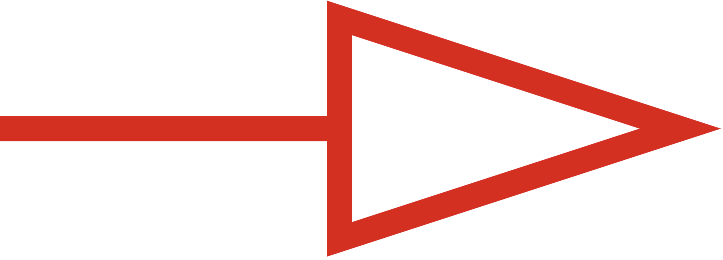}}{B}).
        However, in in HotpotQA-FLM (\protect\scalerel*{\includegraphics{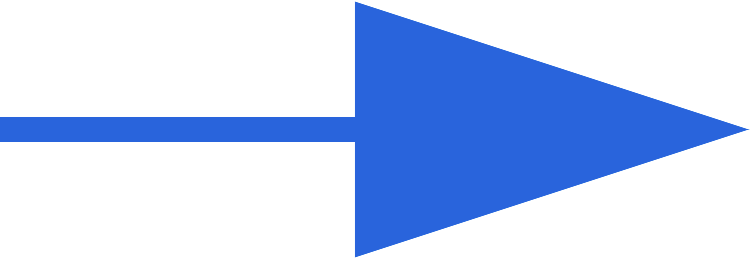}}{B}), critical information is missing \rblue{\textcircled{\raisebox{-0.5pt}{\scriptsize{1}}}}.
        To acquire that information, the ACQ model \rblue{\textcircled{\raisebox{-0.5pt}{\scriptsize{2}}}} first uses the context to generate a clarification question.
        The question is presented to the contextually knowledgeable answering agent \rblue{\textcircled{\raisebox{-0.5pt}{\scriptsize{3}}}}, which generates a response.
        The response is sent as additional context to the downstream model \rblue{\textcircled{\raisebox{-0.5pt}{\scriptsize{4}}}}.
        For strong ACQ models, we expect the downstream model to achieve better performance on context + answering agent response than on context alone.
    }
    \label{fig:general}
\end{figure}
Other tasks such as those by \citet{rao2019answer} use human evaluators to judge the quality and informativeness of questions.
Though human annotators enable flexible and nuanced evaluation, they add a significant cost to the analysis and make results difficult to compare across setups.
These drawbacks make human annotation impractical for large-scale language model benchmarking in the style of BIG-bench \cite{srivastava2022beyond} and MMLU \cite{hendrycks2020measuring}.

Recently, some ACQ tasks including those by \citet{zhang2023clarify} avoid the limitations of word overlap and human evaluations by measuring the effect of clarifications on a downstream task.
In this paradigm, which we refer to as pragmatic evaluation, an answering agent is used to dynamically generate answers to clarifying questions (Figure \ref{fig:general}).
The downstream task, e.g. QA, is then performed with and without the clarification.
Typically both the downstream agent and answering agent roles are performed by language models.
Pragmatic evaluation captures the objective value of the information gained through clarification while also permitting automatic evaluation.

To our knowledge, existing pragmatic asking clarifying question (PACQ) tasks are limited to disambiguating user intent rather than seeking out objective factual information. 
Unfortunately, although largely unexplored, failing to clarify basic facts in high-stakes applications can cause serious harm to users and others.
If a user asks how to clean up a chemical spill, clarifying what chemical is critically important;
applying water to an alkali metal can cause explosion, but sweeping up fine powders can aerosolize toxins. Absorbing oxidizers with paper towels, however, may cause spontaneous combustion \cite{acs2024guide}.
Analogous scenarios exist in medical, legal, security, or other domains where failing to clarify can have serious real-world consequences.
Although our contributions address one specific scope, we find that current models struggle to clarify key facts even in this constrained trivia QA domain.
This suggests more work is necessary before models can adapt to under-specified high-stakes environments.


Compared to ambiguity in user intent, ambiguity in relevant facts poses unique challenges.
Users can generally answer questions about their own intent.
Complex factual questions, however, may be difficult to answer even if unambiguous (Exactly how much did you earn last year?).
Factual questions should be phrased to require minimal effort in responding while still learning facts relevant to the downstream task (Did you earn more than \$X?)

Additionally, when evaluating clarification questions in the factual domain, one must ensure critical pieces of the puzzle are not guessable or leaked in some other way. 
Otherwise, evaluations will inflate the value of gained information.
A task that nominally requires clarification ("Napoleon Bonaparte was 167cm. Who is taller, Shaquille O'Neal or Napoleon Bonaparte?") becomes trivial if the downstream agent is aware that Shaquille O'Neal was a very tall basketball player.


To bridge this gap, we introduce the first PACQ task that focuses on evaluating models' ability to ask questions seeking objective factual information.
Our first contribution is HotpotQA-FLM.
In this task, an LLM must assist a downstream agent in answering a trivia question that is conditional on an unknown fact.
The LLM must identify what information is missing, and ask for it from a third answering agent.
HotpotQA-FLM prompts are created by deleting one fact from the context necessary to perform a downstream multi-hop QA task from the HotpotQA dataset \cite{yang2018hotpotqa}. 
We term this process fact-level masking (FLM).
Clarifying questions are submitted to an answering agent.
The answering agent responds with one of many topically similar answers.
Last, performance on the downstream task is assessed with and without the clarification.

We find state-of-the-art models struggle with HotpotQA-FLM as compared to humans. 
Questions by GPT-4 recover only 53\% of missing information compared to those by humans.
Smaller, open source models achieve only 14\% of human performance.

Given weak zero-shot performance on HotpotQA-FLM, we also contribute a method for training models to ask informative clarification questions.
Like most real-world situations, HotpotQA-FLM does not include examples of clarifying questions for supervised fine-tuning.
Instead, we train our model, Alexpaca, by creating a synthetic dataset through repeated interaction with the answering agent. 
The dataset is filtered with rejection sampling to only include clarifying question examples that result in the expected useful response.
Last, Alexpaca is fine-tuned on the synthetic dataset.
Alexpaca shows a 28\% increase in performance over its zero-shot Llama 3 8B Instruct source model on the full dataset.
This demonstrates small models' ability to self improve at clarifying question generation given effective feedback.
Alexpaca also demonstrates a scalable and cheap proof-of-concept for approaching factual ACQ tasks.
The training method is suitable where supervised examples are unavailable or proprietary models are unacceptable for cost, privacy, or latency reasons.

To summarize, our contributions are: 1) HotpotQA-FLM, a clarification question generation task evaluated based on objective information gain in the factual domain, and 2) Alexpaca, a rejection-sampling approach to fine-tuning models for clarification question generation not reliant on manual annotation.

\section{Related Work}

\subsection{General Question Generation}

Question Generation (QG), speaking generally, is the task of automatically generating questions \citep{rus2008question}.
Questions can be generated using syntactic \citep{gates2008generating,yao2012semantics} or neural \citep{chen2018learningq} approaches.
\citet{duan2017question} and \citet{wang2020multi} generate questions for data augmentation for QA tasks and pretraining, respectively, using convolutional, recurrent, and transformer architectures.
Chatbots designed for social dialogue may ask questions to exhibit emotional intelligence, prompt users, and drive engagement \citep{shum2018eliza}.
Question-asking can also be used for educational purposes \citep{kurdi2020systematic}.
Four automatically evaluated question generation tasks appear in BIG-bench \cite{srivastava2022beyond} including Twenty Questions, Forecasting Subquestions, Question-Answer Generation, and Question Selection.

\subsection{Asking Clarifying Questions}
Asking clarifying questions (ACQ) is a type of QG for acquiring additional factual knowledge or disambiguating user intent, as in \citep{aliannejadi2019asking}. 
During general QG, outputs are often evaluated based on the Bleu, Rouge, or other word overlap metrics, as in \citep{qi2020stay}, \citep{xu2019asking}, \citep{min2020ambigqa}, \citep{deng2022pacific}, \citep{gaur2022iseeq}, \citep{chen2018learningq}, \citep{meng2023followupqg} \citep{kostric2024generating} \citep{ang2023socratic}. 
Other research uses human evaluations, \citep{pyatkin2022clarifydelphi}, \citep{rao2019answer}, \citep{rao2018learning}, \citep{chen2022generating}.
Pragmatic asking clarifying questions (PACQ), on the other hand, evaluates a question based on the usefulness of the answer it prompts (Figure \ref{fig:general}).
\citep{zhang2023clarify}, \citep{lee2023asking} and \citep{andukuri2024star} explore ACQ pragmatically but in the intent rather than factual domain.
The GuessWhat?! \citep{de2017guesswhat} and CLEVR Ask \citep{matsumori2021unified} explore constrained iterative binary PACQ tasks in the vision domain.

\subsection{Related Tasks}

In task-oriented dialog (TOD), the system is designed to converse with the user to perform a slot-filling task.
Slot-filling tasks are typically straightforward and well-defined, like booking a hotel.
What information is missing, such as the desired price range, is usually easily defined by which slots are empty \citep{budzianowski2018multiwoz}.
In such TOD cases, template-based systems can be sufficient, with the main challenge being natural language understanding and dialog state tracking.
Since the set of useful questions is neither complex nor numerous, TOD systems often assume that the user will be able to answer all system-generated questions.
By decoupling TOD from a fixed slot ontology and accounting for incomplete user knowledge, PACQ can be viewed as a generalization of the dialog planning and natural language generation steps of TOD.

Finally, PACQ is similar to the idea of tool-use, where models can consult APIs like a calculator, search engine, or QA model to improve performance on a downstream task.
Tool-use models like Toolformer \citep{schick2023toolformer} call APIs internally during generation to gather additional knowledge.
Framing PACQ as a distinct task may improve data efficiency in training and granularity of evaluation as compared to end-to-end tool use.

\section{Methods}

\subsection{Problem Description}

The goal of pragmatic asking of clarifying questions is for the ACQ model to transfer information from a knowledgeable answering agent to an executive downstream model by asking a clarifying question. 
In our setup the answering agent is a language model, but could also be a database, human expert, or the user.
The downstream model is a model that directly executes some task for the user, such as a legal assistant chatbot or QA model. 
The answering agent is an agent capable of answering clarification questions related to the downstream task.
This could be a human user, expert, or LLM stand-in like Flan-T5 \citep{chung2022scaling}. 
The ACQ model is a language model agent capable of generating questions that assist the downstream model in its task. 
It takes the downstream task as input and generates a question for the answering agent. 
The answering agent response is concatenated to the original context and then passed to the downstream model, giving the downstream model access to the information requested in the question. 
The ACQ model's performance is evaluated using the difference between the downstream model's performance with and without the answering agent's answer.

Many factors affect the extent of PACQ performance gains elicited by answering agent's responses, including the context, the bias of the models, the possible responses, and what information has been memorized by the downstream model.
Hence, PACQ performance can only be assessed in the context of a particular answering and downstream agent.
Our setup, as described above and similar to \citep{lee2023asking}, consists of a downstream model, $D$, tasked with performing some task, and an answering agent, $A$, which responds to questions generated by the ACQ model, $C$.
In the next section, we present a specific $C \rightarrow A \rightarrow D$ setup and dataset on which to evaluate it.

\subsection{Model Training}

Creating examples of good clarification questions is expensive and challenging because question usefulness depend on the properties of the answering and downstream agents.
Any change to these agents may require a different question generation strategy.
Therefore, it is preferable that models be trained through interaction with the answering agent rather than through manual supervision.
We propose one method for performing such training.
In our method, we use a zero-shot model to repeatedly generate clarifying questions.
We then fine-tune the zero-shot model on only the clarifying questions that produce useful information.

\subsection{Problem Definition}

Let $t$ be a natural language statement of a task.
Let $f_1, ... , f_n$ be natural language facts consisting the context for the task.
Let example $x=t+f_{1}+...+f_{n}$, where $+$ indicates string concatenation and $-$ will represent deletion.
Let $D(x) \rightarrow y$ be a downstream model that takes $x$ as input and outputs $y$. 
Let $C(x) \rightarrow q$ be a ACQ model that takes $x$ as input and generates a natural language question $q$. 
Let $R(D,x,y) \rightarrow r$ be some reward on which $D$ is evaluated, where more positive values are better, such as F-score, accuracy, or negative loss.
For brevity, we often omit $D$ and $y$.

We say a fact $f$ is supporting if it is believed that $R(x+f) > R(x - f)$. Otherwise we say $f$ is distracting \citep{yang2018hotpotqa}.
Let $A(q) \rightarrow f_r$ be an answering agent that takes $q$ as input and returns a response $f_r$.
The PACQ task is to create a model $C$ that maximizes
\[
\begin{aligned}
    \Delta r    & = R(x+f_r) - R(x)
\end{aligned}
\]
One may construct more complex versions of PACQ involving multiple missing facts, iterative asking, multiple answering agents, or cost functions for different types of questions. 
In this paper, we limit PACQ to the costless, single-mask, single-turn, single-answering agent case and we do not address determining whether a task lacks sufficient context.

\section{Experiments}
\subsection{Dataset}
We contribute FLM-HotpotQA, a version of the QA dataset HotpotQA for evaluating pragmatic asking of clarifying questions \citep{yang2018hotpotqa}.
HotpotQA is a multi-hop QA reasoning task where each example contains both supporting and distractor facts from Wikipedia as determined by human annotators.
We choose reward function $R$ to be the F1 score of the word overlap between the predicted answer and the ground truth answer following the original HotpotQA. 
Thus $r \in [0,1]$ and $\Delta r \in [-1, 1]$.

\begin{figure*}
    \centering
    \begin{tikzpicture}
        \begin{small}
            \draw[rounded corners, BrickRed] (-7.2,.5) rectangle (8.0, 2.2);
            \node[anchor=south east, BrickRed] at (7.6, .4) {Incomplete Example $x^i$};
            \draw[rounded corners, RoyalBlue] (-7.0,-.4) rectangle (7.8, 2.4);
            \node[anchor=south east, RoyalBlue] at (7.6, -.5) {Complete Example $x^c$};
            \draw[rounded corners, Green] (-6.8,-2.2) rectangle (7.6, 1.7);
            \node[anchor=south east, Green] at (7.6, -2.3) {Candidate Oracle Responses};
            \end{small}
        \node [draw=none] (example-tabular) {
            \setlength{\tabcolsep}{2pt} 
            \renewcommand{\arraystretch}{1.5}
            \begin{small}
                \begin{tabular}{l p{.67\textwidth}}
                    t & When was the composer of "Persian Surgery Dervishes" born?\\
                    $f^{sup}_1$ & Persian Surgery Dervishes is a recording of two live solo electric organ concerts, the first held in Los Angeles on 18 April 1971 and the second in Paris on 24 May 1972, by avant-garde minimalist composer Terry Riley.\\
                    $f^{sup}_2$ ($f^*$) & Terrence Mitchell "Terry" Riley (born June 24, 1935) is an American composer and performing musician associated with the minimalist school of Western classical music.\\
                    $f^{dis}_1$ & Thomas Christian David (December 22, 1925 - January 19, 2006) was an Austrian composer, conductor, choral conductor, and flutist. \\
                    $f^{dis}_2$ & Abdolreza Razmjoo is a composer, arranger and singer Tenor of Iran Kurdish ancestry from Kermansha. \\
                \end{tabular}
            \end{small}
        };
    \end{tikzpicture}
    \caption{An example containing a downstream task t, supporting facts $f^{sup}_{1,...,n}$, and distractor facts $f^{dis}_{1...n}$. 
    (Additional facts not shown.)
    We create an \textcolor{BrickRed}{incomplete example $x^i$} by masking one supporting fact, $f^*$, chosen at random, from the facts in the \textcolor{RoyalBlue}{complete example $x^c$}.
    Prompted with $x^i$, the ACQ model poses a question to the answering agent which returns one \textcolor{Green}{answering agent response $f_r$} from the supporting or distractor facts. We then append $x^r=x^i+f_r$, which we expect to improve downstream model performance $D(\cdot)$
    }
    \label{fig:example}
\end{figure*}

To evaluate our our ACQ model, we will create three context examples: the incomplete example $x^i$ missing some context, the complete example $x^c$ with full context, and $x^r$ which contains the incomplete context plus additional context derived from the clarifying question. 
The incomplete and complete contexts will serve as the worst- and best-case benchmarks against which we compare the response context.

First construct $x^c$ that contains the task and every supporting fact (Figure \ref{fig:example}).
Next, we apply fact-level masking to each example.
From each complete example, we create an incomplete example $x^i$ by randomly selecting one supporting fact, $f^*$, to be the masked fact and deleting it from the context: $x^i = x^c - f^*$.
When missing one supporting fact, the downstream task becomes substantially more difficult, even for strong zero-shot models like GPT-4 \cite{openai2023gpt4} (Figure \ref{fig:m1_ablation}).
The masked fact, along with the distractor facts and the other supporting facts, make up the set of responses, $f_r$, the answering agent may give.
Finally, we prompt the question model with the incomplete context to generate a question, then generate a response $f_r$ from the answering agent. We create the response example $x^r$ by appending $x^r = x^i + f_r$.
To benchmark human performance, one author of this paper annotated a test set of 400 clarifying questions from examples also included in the full set.
    
In general, we expect the complete example $x^c$, which contains every supporting fact, to have the highest possible reward.
Meanwhile, we say an example $x$ is improvable if there exists at least one possible response $f_r$ such that $\Delta r(f_r) > 0$.
By masking facts in $x^c$ we can decrease the reward on the example, producing an improvable self-supervised example.
Note that not all incomplete examples will be improvable, for example, if: 
\begin{itemize}[topsep=0pt,itemsep=-1ex,partopsep=1ex,parsep=1ex]
    \item Two facts contain redundant information
    \item $D$ has memorized knowledge of information in $f^*$
    \item $f^*$ is mislabeled as supporting
    \item $x^i$ still allows $D$ to make a spurious correlation without $f^*$
\end{itemize}
It is also possible for $x^i$ to be improved by a response $f_r$ even if $f_r \neq f^*$, if $f_r$ and $f^*$ contain similar information.
27.6\% and 28.5\% of examples in our Full and test sets, respectively, are improvable.
We preserve unimprovable examples in the dataset to avoid bias;
the downstream model may sometimes achieve the correct response through a spurious correlation on the incomplete example, but fail to make the spurious correlation after recieving the response.
Similarly, the downstream model may fail even given the masked fact, but succeed given another fact if the other fact contains more helpful information. 

\subsection{Evaluation Implementation Details}
To generate and evaluate answers to PACQ questions, we construct the following pipeline.
The ACQ model $C$ takes an incomplete example $x^i$ as input to generate a clarifying question $q$. 
As baselines for $C$ we choose GPT-4 \citep{openai2023gpt4}, ChatGPT \citep{chatgpt}, Llama 3 8B Instruct \citep{llama3modelcard}.
We select these models for their strong performance on zero-shot tasks.
We choose a prompt template for each model by evaluating three zero-shot and three 5-shot in-context prompts on 400 examples from the training dataset.
In addition, we create a new model, Alexpaca, by fine tuning Llama 3 on a dataset of its own generations filtered with rejection sampling.
Finally, we include a dummy Repeater model among the baselines, which simply returns the input task.

Questions generated by $C$ are passed to the answering agent $A$, a Flan-T5-Base model, which we choose for its accessibility and strong zero-shot performance on other QA tasks. 
The answering agent serves as a stand-in for a human expert answering clarifying questions generated by $C$.
$A$ returns $f_r$, the most likely response to $q$ from among all possible distractor facts $F^{dis}$ present in the original HotpotQA example ($\text{mean}=39.2$, $\text{std}=11.4$), all supporting $F^{sup}$ facts ($n-1$ of which are already present in the context, $\text{mean}=1.43, \text{std}=0.71$), and the masked fact $f^*$. 

To create the response example $x^r$, we append the answering agent response to the incomplete example.
Note that by appending rather than inserting, the order of facts may be altered as compared to $x^c$, even if $f_r=f^*$, which may occasionally affect the output of the downstream model. 

Finally, we compare downstream model, $D$, performance given $x^i$, $x^r$, and $x^c$. 
$D$ is also a Flan-T5-Base model. 
We choose Flan-T5-Base over models using more parameters or training data because we expect they are more likely to answer based off of context rather than information memorized from training data (e.g., Wikipedia).
If $C$ produces a question with positive utility towards $D$, then one should expect $R(x^c) \geq R(x^r) > R(x^i)$.
To express reward relative to its theoretical minimum ($R(x^i)$) and maximum ($R(x^c)$) values, we define recovery as:
\begin{align*}
    \rho = 100 \cdot \frac{R(x^r)-R(x^i)}{R(x^c)-R(x^i)}
\end{align*}
and select F1 recovery as our downstream evaluation metric.

\subsection{Alexpaca: Fine-Tune through Interaction}

Annotating high quality clarifying questions is challenging and costly.
For this reason, we train our model, Alexpaca, purely through interacting with the answering agent.
First, we the Llama 3 8B Instruct foundational model to generate a set of clarifying question examples using rejection sampling.
To ensure examples are of high quality, we reject questions if the answering agent response does not match the masked fact. 
We repeat the generation for each example until one is accepted, or until $k=40$ rounds. 
Each round we increase generation temperature by $2/k$, starting at 0.01 in order to encourage exploration in later rounds.
Finally, we fine-tune the same Llama 3 foundational model on the rejection sampling dataset.

\section{Results and Discussion}

\subsection{Baseline Performance}

\begin{figure}
    \begin{minipage}[]{0.49\textwidth}
        \centering
        \includegraphics[width=\textwidth]{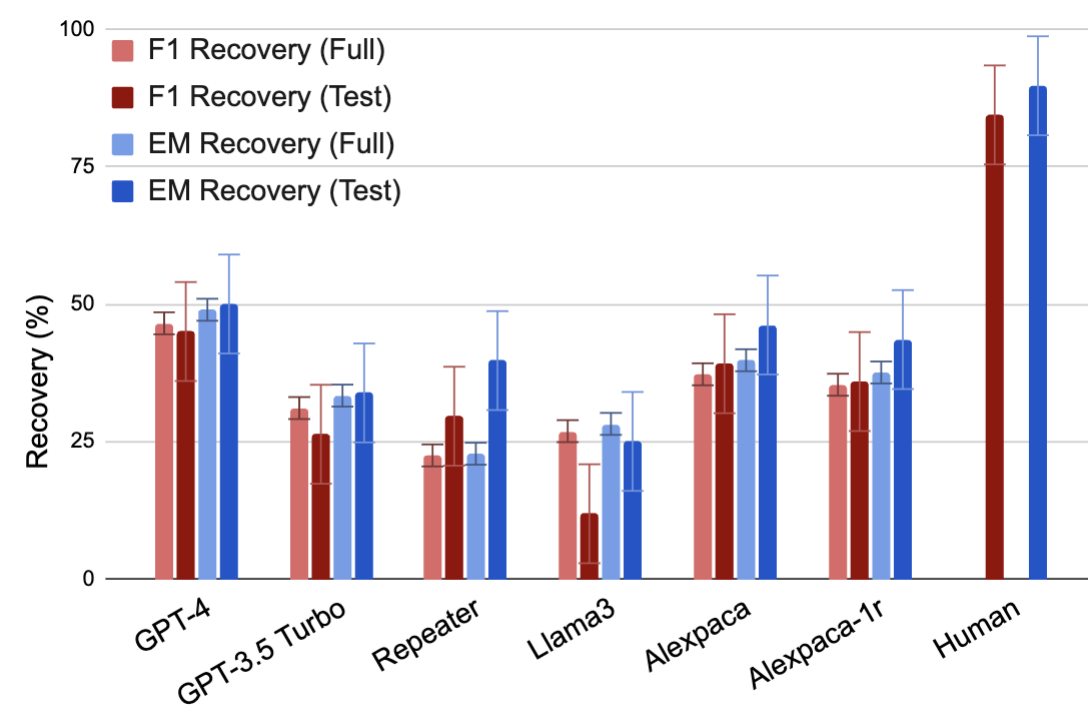}
        \caption{F1 and exact match recovery for PACQ models and human annotators. Results shown for the Full validation set ($n=7404$) and the test set ($n=400$), which contains human-generated ACQ questions. Alexpaca-1r indicates single round rejection sampling.}
        \label{fig:main}
    \end{minipage}
\end{figure}

We report F1 and exact match recovery results for ACQ models on the full HotpotQA validation set ($n=7404$, Figure \ref{fig:main}). 
Of all models, GPT-4 performs best in both F1 and exact match (EM), recovering 46.5\% and 49.0\% respectively.
These results, however, fall short of complete recovery of missing information, indicating room for improvement even in strong zero-shot models.
Other models perform substantially worse.
Llama 3 achieves 26.9\% F1 recovery, which is only a moderate improvement over the dummy Repeater model. We suspect Repeater achieves its positive recovery (22.5\%) by exploiting a bias in the answering agent towards choosing responses with high keyword overlap with the input question.

\subsection{Alexpaca Fine-Tuning Performance}

Alexpaca exceeds baseline Llama 3 performance by 37.2\% vs. 26.9 F1 recovery ($p=0.00074$), demonstrating a method for self-improving ACQ models given an answering agent rather than example clarifying questions. 
We report the average of results for five random seeds.
During training dataset creation, repeatedly attempting to generate passing examples up to 40 times each (Alexpaca) improves F1 recovery by 6.0\% points compared to using a single attempt (Alexpaca-1r).
We believe that challenging examples accepted in later rounds of rejection sampling and generated at higher temperature have a disproportionate effect on model behavior.

\subsection{Alexpaca Behavior}

Although Alexpaca elicits the masked fact more often than GPT-4 on the test set (189 vs. 162), Alexpaca's overall improvement rate is still lower (72 vs. 79). 
Likely this is an artifact of the Alexpaca training rejection criteria wherein acceptance is determined by eliciting the masked fact rather than actual downstream improvement. 
This indicates room for improvement in baseline models performing PACQ.
Attempts to correct this bias by accepting examples based on recovery rather than masked fact response did not achieve statistically significant improvement in F1 recovery, possibly due to a higher signal-to-noise ratio in end-to-end systems.
Although GPT-4 achieves higher performance than Alexpaca, Alexpaca is open-source and uses many times fewer parameters compared to GPT-4.
Furthermore, Alexpaca outperforms GPT-3.5 Turbo in all metrics.
This makes Alexpaca more suitable in circumstances where cost, latency, or privacy are a concern.

\begin{figure}
    \begin{minipage}[]{0.49\textwidth}
        \centering
        \begin{subfigure}[]{0.49\textwidth}  
            \centering 
            \includegraphics[width=\textwidth]{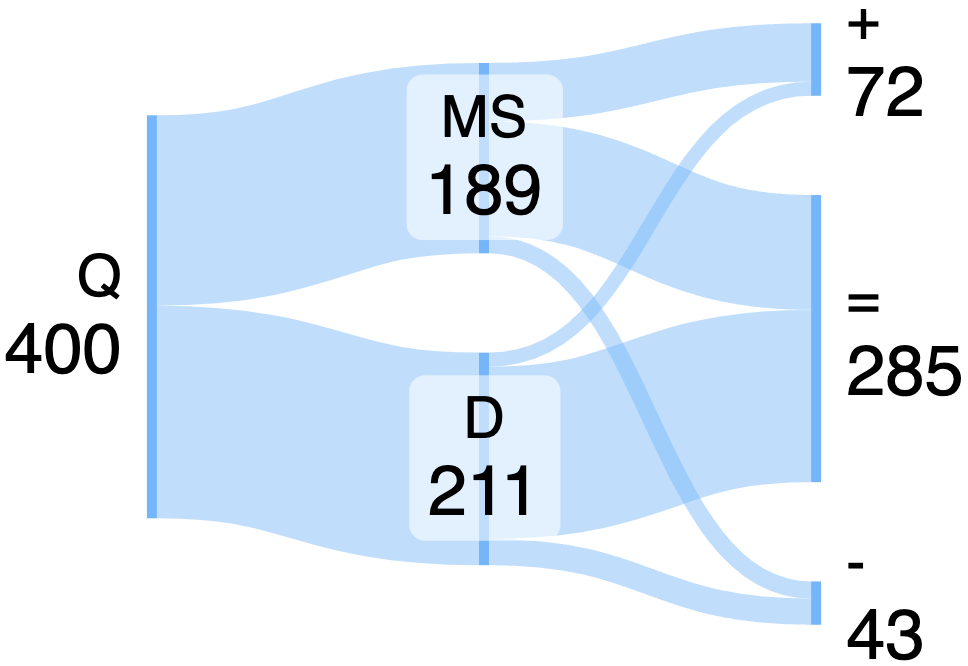}
            \caption[]%
            {{\small Alexpaca}}    
            \label{fig:sankey_alexpaca}
        \end{subfigure}
        \hfill
        \begin{subfigure}[]{0.49\textwidth}
            \centering
            \includegraphics[width=\textwidth]{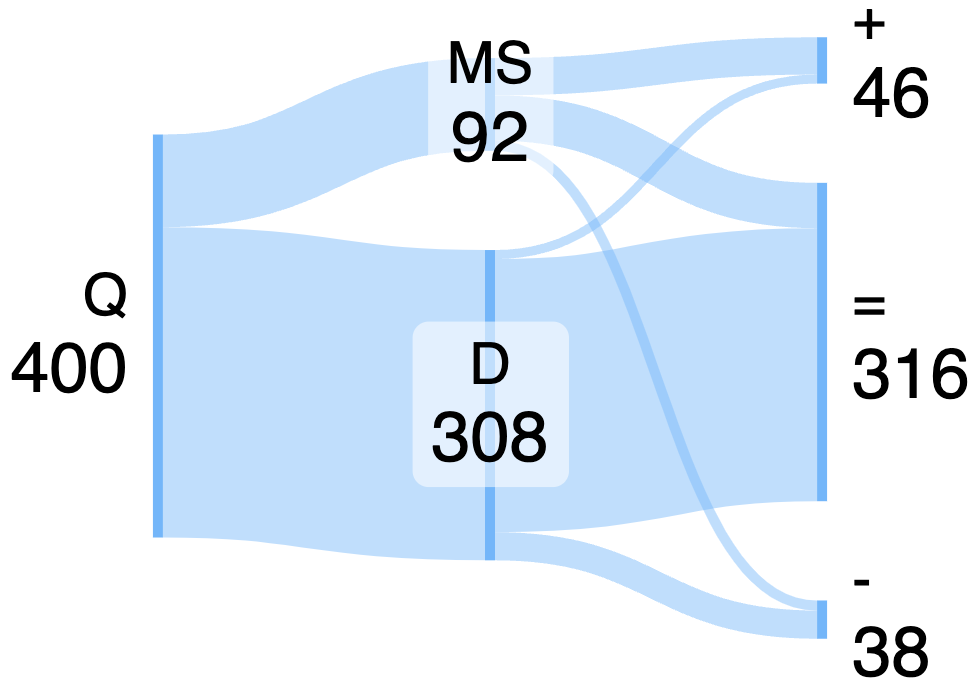}
            \caption[]%
            {{\small Llama 3}}    
            \label{fig:sankey_chat}
        \end{subfigure}
        \begin{subfigure}[]{0.49\textwidth}   
            \centering 
            \includegraphics[width=\textwidth]{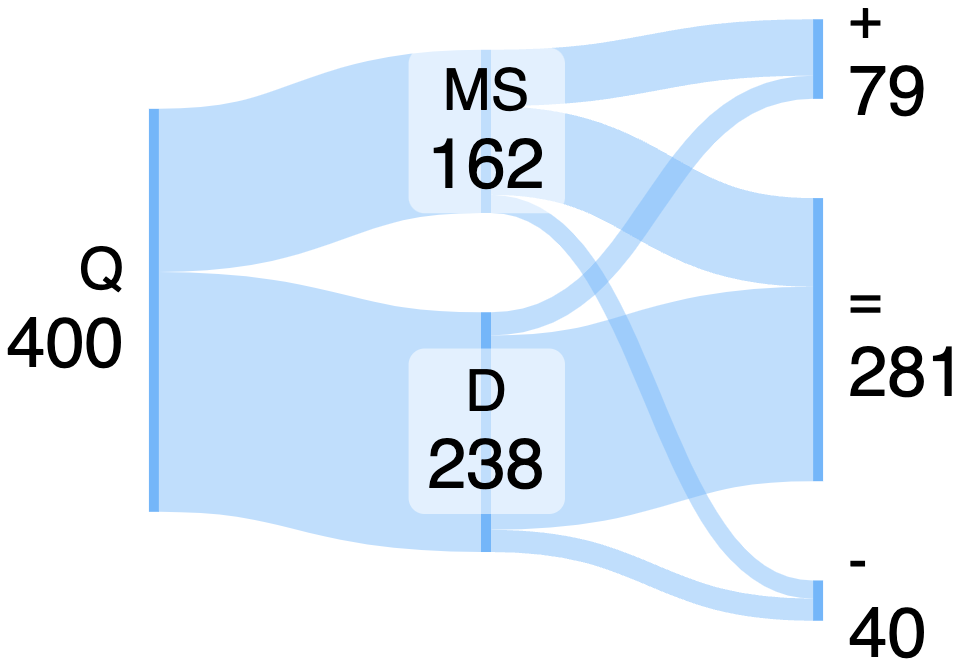}
            \caption[]%
            {{\small GPT-4}}    
            \label{fig:sankey_repeater}
        \end{subfigure}
        \begin{subfigure}[]{0.49\textwidth}   
            \centering 
            \includegraphics[width=\textwidth]{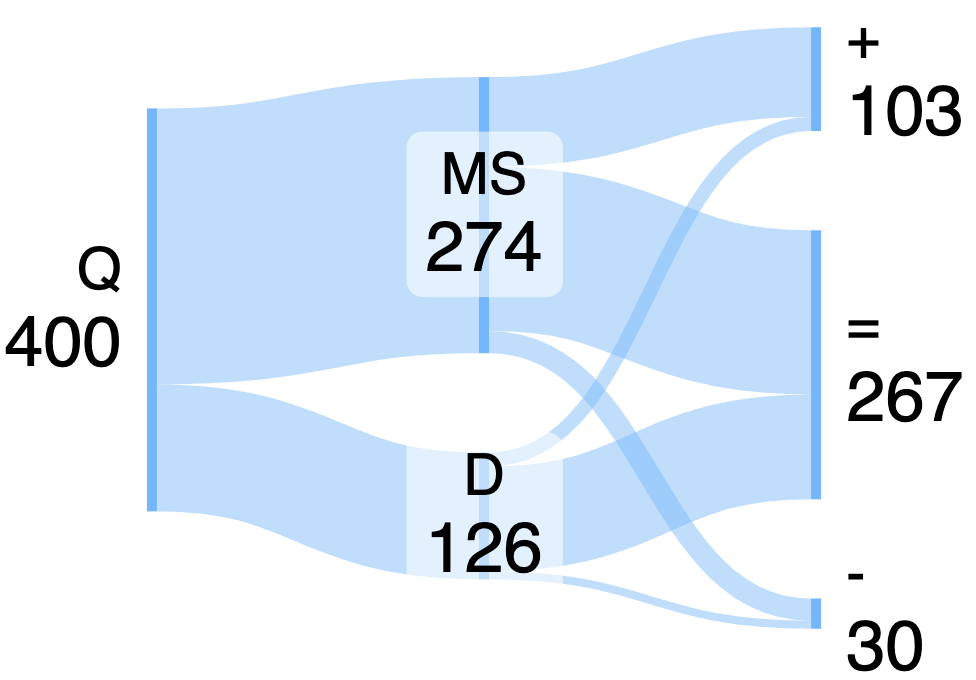}
            \caption[]%
            {{\small Human Annotation}}    
            \label{fig:sankey_gt}
        \end{subfigure}
        \caption[ The average and standard deviation of critical parameters ]
        {Proportion of questions (Q) answered with a masked fact (MS) vs. distractor (D) by answering agent (left section). Proportion of answers given resulting in positive, zero, or negative difference in downstream model performance (right section).}
        \label{fig:sankey}
    \end{minipage}
\end{figure}

\subsection{Comparison to Human Performance}

We find that human-generated questions on the test set are more likely to elicit the masked fact $f^*$ in the response (Figure \ref{fig:sankey}).
Eliciting the masked sentence usually, but not always, produces as good or better a result in the downstream model compared to eliciting a distractor.
This leads to human annotations performing significantly better than the best baseline models.
Human annotation achieved 84.4\% F1 and 89.7\% EM recovery, compared to the strongest baseline, GPT-4, which achieved 45.0\% F1 and 50.0\% EM recovery on the test set (Figure \ref{fig:main}).

\begin{table*}
    \small
    \centering
    \begin{tabularx}{1.0\textwidth}{lcccccccccccc} 
        \cmidrule{1-12}
        &\multicolumn{4}{c}{Full} & &\multicolumn{4}{c}{Test} \\
        \cmidrule{2-6}\cmidrule{8-12}
    Model &F1 &F1 Recovery &EM &EM Recovery &MFRR & &F1 &F1 Recovery &EM &EM Recovery &MFRR \\
            \cmidrule{1-12}
    GPT-4 &62.5 &46.5 &46.8 &49.0 &39.3 & &61.4 &45.0 &47.5 &50.0 &40.5 \\
    GPT-3.5 Turbo &59.8 &31.1 &44.6 &33.3 &25.7 & &57.9 &26.3 &44.8 &33.8 &22.8 \\
    Repeater &58.3 &22.5 &43.1 &22.8 &29.1 & &58.5 &29.6 &45.8 &39.7 &32.8 \\
    Llama 3 &59.1 &26.9 &43.9 &28.2 &22.8 & &55.2 &11.8 &43.3 &25.0 &23.0 \\
    Alexpaca &60.9 &37.2 &45.5 &39.8 &39.2 & &60.3 &39.1 &46.9 &46.2 &45.3 \\
    Alexpaca-1r &60.5 &35.3 &45.2 &37.5 &37.9 & &59.7 &35.9 &46.4 &43.5 &43.1 \\
    Human &- &- &- &- &- & &68.8 &84.4 &54.3 &89.7 &68.5 \\

        \cmidrule{1-12}
        
    \end{tabularx}
    \caption{F1 and exact match recovery for full validation set and manually annotated test set.}\label{tab:main}
\end{table*}

\subsection{Downstream Model Ablation}
\begin{figure}
    \begin{minipage}[t]{0.49\textwidth}
        \centering
        \includegraphics[width=\textwidth]{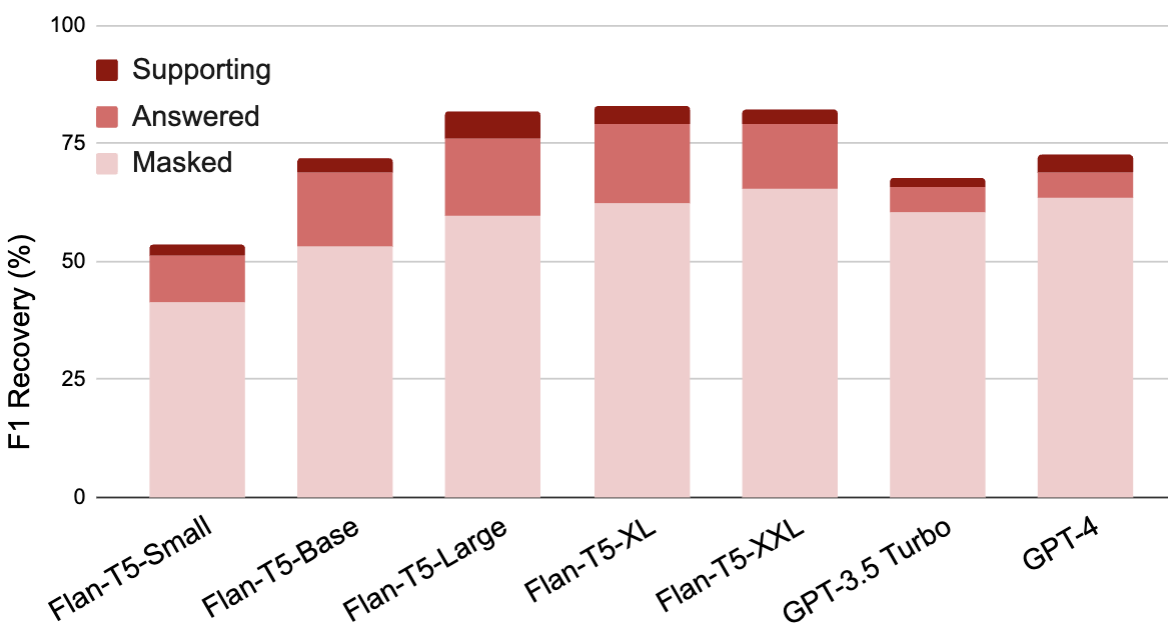}
        \caption{Supporting, answered, and masked F1 as a function of downstream model architecture.}
        \label{fig:m1_ablation}
    \end{minipage}
\end{figure}

We evaluate all available sizes of Flan-T5, GPT-3.5 Turbo, and GPT-4 as candidate downstream models using a Flan-T5-Base model as the answering agent and human-generated questions as the ACQ model.
Models lose between 7.4\% (GPT-3.5 Turbo) and 22.0\% (Flan-T5-Large) absolute points F1 score as a result of masking a single supporting fact (Figure \ref{fig:m1_ablation}).
Models recover between 59.1\% (GPT-3.5 Turbo) and 84.4\% (Flan-T5-Base) of the F1 score lost during masking after including the answering agent response to human generated questions.
Although models are affected differently by FLM, with GPT-X models being more robust, reasonable consistency in F1 recovery rate suggests that the choice of downstream model has minimal impact on PACQ evaluation.
We suspect GPT-X models are more robust than Flan-T5 since in exploration they appear to have memorized large portions of Wikipedia, which minimizes the impact of removing Wikipedia facts from context.

\subsection{Answering Agent Ablation}

\begin{figure}
    \hfill
    \begin{minipage}[t]{0.49\textwidth}
        \centering
        \includegraphics[width=\textwidth]{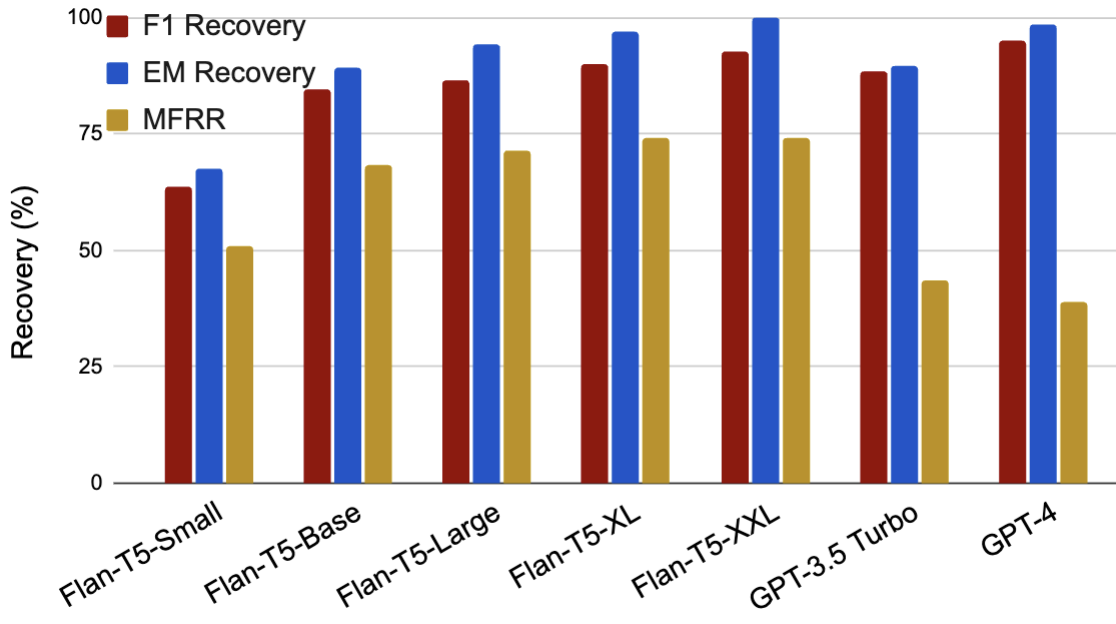}
        \caption{F1, exact match and masked fact response rate (MFRR) as a function of answering agent size and architecture.}
        \label{fig:oracle_ablation}
    \end{minipage}
\end{figure}

We test GPT-3.5 Turbo, GPT-4, and all sizes of Flan-T5 as the Answering Agent on human-generated questions.
Flan-T5-Base and larger respond with the masked fact in more than 68\% of cases (Figure \ref{fig:oracle_ablation}). 
Furthermore, we observe consistently strong performance by these models on F1 and exact match, with both metrics exceeding 84\% recovery in all cases. 
This indicates that when prompted by well-formed and informative questions, Flan-T5 of size Base and larger can consistently respond with appropriate answers.
For the sake of accessibility, we choose the smallest strong model, Flan-T5-Base, as our answering agent.
Interestingly, although GPT-4 responds with the masked fact far less frequently than any Flan-T5 model (GPT-4: 43.5\%, Flan-T5-XXL: 74.0\%), GPT-4 achieves the highest F1 recovery overall and 98.5\% exact match recovery. 
This suggests that although GPT-4 gives distractor or redundant supporting facts most of the time, the facts it chooses still carry critical information.
This illustrates the importance of measuring information gain rather than nominal correctness.

\subsection{Error Analysis}

We observe one failure mode associated with the answering agent and three associated with the ACQ model, which prevent PACQ questions from recovering missing information.
Most obviously, the answering agent may return an irrelevant and unhelpful response.
In 31.5\% of cases, human-generated questions induce responses other than the masked fact.
When $f^* \neq f_r$, the F1 score of the downstream model increases in only 11.1\% of cases, compared to 32.5\% of cases when $f^*=f_r$ (Figure \ref{fig:sankey}d).
When a distractor fact does cause an increase in F1, it is often because information in the distractor fact contains overlaps with information in the masked fact.

Other times, the failure mode is due to the ACQ model generating poor questions. 
In some examples, GPT-4 asks for information already present in the context:

\begin{tcolorbox}[colback=blue!2!white,colframe=RoyalBlue!90!gray,left=2pt,right=2pt,top=2pt,bottom=2pt]
\footnotesize
\noindent \textbf{Context}: The 1925 Birthday Honours were appointments by King George V to various orders and honours to reward and highlight good works by citizens of the British Empire.

\noindent \textbf{Task}: In which year was the King who made the 1925 Birthday Honours born?

\noindent  \textbf{GPT-4}: Who was the King that made the 1925 Birthday Honours? \bred{\ding{55}}

\noindent  \textbf{Human}: When was George V born? \fgreen{\ding{51}}
\end{tcolorbox}

In other cases, GPT-4 appears to struggle with simple categorical reasoning:

\begin{tcolorbox}[colback=blue!2!white,colframe=RoyalBlue!90!gray,left=2pt,right=2pt,top=2pt,bottom=2pt]
\footnotesize
\noindent \textbf{Context}: Backford Cross is located around the A41/A5117 junction, south of Great Sutton and about 1.5 mi north of the village of Backford, near Chester.

\noindent \textbf{Task}: What major truck road is located in Backford Cross?

\noindent  \textbf{GPT-4}: What major roads are near Backford Cross? \bred{\ding{55}}

\noindent  \textbf{Human}: Is the A41 or the A5117 a major trunk road? \fgreen{\ding{51}}
\end{tcolorbox}

Sometimes GPT-4 generates entirely useless questions, such as "Who was president during the Nixon administration?" (Richard Nixon).

Alexpaca makes mistakes similar to GPT-4. Additionally, Alexpaca may hallucinate relevant details to ask about:

\begin{tcolorbox}[colback=blue!2!white,colframe=RoyalBlue!90!gray,left=2pt,right=2pt,top=2pt,bottom=2pt]
\footnotesize
\noindent \textbf{Context}: Dinosaur is a 2000 American CGI animated adventure film produced by Walt Disney Feature Animation and The Secret Lab and released by Walt Disney Pictures.

\noindent \textbf{Task}: Which came out first, Dinosaur or McFarland, USA

\noindent  \textbf{Alexpaca}: What Disney animated film was released in 1942? \bred{\ding{55}}

\noindent  \textbf{GPT-4}: When was McFarland, USA released? \fgreen{\ding{51}}
\end{tcolorbox}

Since the Alexpaca training approach does not focus on style improvement, Alexpaca sometimes retains Llama 3's prototypical cheery chatbot verbosity, though more frequently asks informative questions:

\begin{tcolorbox}[colback=blue!2!white,colframe=RoyalBlue!90!gray,left=2pt,right=2pt,top=2pt,bottom=2pt]
\footnotesize
\noindent \textbf{Context}: Sir Arthur Ignatius Conan Doyle (22 May 1859 – 7 July 1930) was a British writer best known for his detective fiction featuring the character Sherlock Holmes.

\noindent \textbf{Task}: Who was born first, Arthur Conan Doyle or Penelope Lively?

\noindent  \textbf{Llama 3}: A clever question!

To answer the final question, I can ask: "When was Arthur Conan Doyle born?" \bred{\ding{55}}

\noindent  \textbf{Alexpaca}: A clever question!

To answer this question, I can ask: "When was Penelope Lively born?"

This question will help me determine the answer to the original question by allowing me to compare the birth dates of Arthur Conan Doyle (22 May 1859) and Penelope Lively (1947). \fgreen{\ding{51}}
\end{tcolorbox}

Although verbose, Alexpaca's question does in fact produce the masked sentence, also illustrating the answering agent's robustness to noise and style. Ironically, Alexpaca identifies that Penelope Lively's birthdate (17 March 1933) is missing, then hallucinates it to be in 1947. 







\section{Conclusion}

In this paper, we presented a benchmark for objectively evaluating clarifying questions and observed that state-of-the-art zero-shot LLMs struggle at this task compared to humans.
To overcome these challenges, we introduced fact-level masking and FLM-HotpotQA, a self-supervised PACQ dataset, and an associated evaluation process.
Finally, we demonstrated a training method for the Alexpaca model that relies on agent-agent interaction rather than supervised examples of clarifying questions to self-improve over baseline.

One limitation of the Alexpaca approach is that it requires answering agent responses to be labeled as useful or not useful.
The FLM process produces such labels implicitly.
In the real-world, however, whether classifying answers is more practical than annotating clarification questions examples depends on the situation.
Nonetheless, we believe this dataset and approach can be used to improve factual clarification question generation in language models and LLM safety in high-stakes, ambiguous environments.




\bibliography{aaai25}

\clearpage
\newpage
\section{Appendix}

\subsection{List of Prompts}
\label{apx:prompts}
\begin{enumerate}
    \item Ask another question that would help you answer the following question: \{context\} \{q1\}
    \item Some information is missing from this context. Ask a simpler question that would help you answer it. Context: \{context\} Main Question: \{q1\} Simpler question:
    \item What question can you ask to help you answer the final question? \{context\} \{q1\} You can ask:
    \item Ask another question that would help you answer the following question: \{in-context examples\} \{context\} \{q1\}
    \item Some information is missing from this context. Ask a simpler question that would help you answer it. \{in-context examples\} Context: \{context\} Main Question: \{q1\} Simpler question:
    \item What question can you ask to help you answer the final question? \{in-context examples\} \{context\} \{q1\} You can ask:
\end{enumerate}

Based on performance on $n=400$ examples from the HotpotQA train dataset we select prompt 3 for Llama 3, GPT-3.5 Turbo, and GPT-4, though improvement over other prompts was not statistically significant.

\subsection{Answering Agent Implementation Details}

For Flan-T5 answering agents, we prompt the model with \\

\noindent Question: \{clarifying question\}\textbackslash n context: \{candidate answer\}\textbackslash n prompt: Does the context answer the question, yes or no? \\

We then return the answer with the highest ratio of the "yes" to "no" logits. For GPT-X answering agents, we prompt the model with\\

\noindent Question: \{clarifying question\}\textbackslash n \textbackslash n \{enumerated answers\} \textbackslash n\textbackslash n Which answer is correct? Only say the number of the answer, nothing else.\\

and return the answer at the index returned. If no valid index is returned, we return a random answer.

\subsection{Downstream Agent Implementation Details}

For downstream agents, we prompt the model with \\

\noindent \{task\} \{article title 1\}: \{fact 1\} ... \{article title n\}: \{fact n\} Answer in as few words as possible: 

\onecolumn

\begin{minipage}{\textwidth}

    \subsection{Answering Agent Architecture Ablation}
    \begin{table}[H]\centering
        \caption{Answering agent architecture ablation for answering agents using Flan-T5-Base as downstream model on the full validation set.}\label{tab:oracle_ablation}
        \begin{tabular}{lrrrrrr}\toprule
            &F1 &F1 Recovery &EM &EM Recovery &MFRR \\\midrule
            Flan-T5-Small &64.9 &63.8 &50.5 &67.6 &50.8 \\
            Flan-T5-Base &68.8 &84.4 &54.2 &89.4 &68.5 \\
            Flan-T5-Large &69.2 &86.5 &55.0 &94.1 &71.3 \\
            Flan-T5-XL &69.8 &90.1 &55.5 &97.1 &74.3 \\
            Flan-T5-XXL &70.4 &92.9 &56.0 &100.0 &74.0 \\
            GPT-3.5 Turbo &69.5 &88.4 &54.3 &89.7 &43.5 \\
            Incomplete &53.0 &0.0 &39.0 &0.0 &- \\
            Complete &71.7 &100.0 &56.0 &100.0 &- \\
            \bottomrule
        \end{tabular}
    \end{table}
\end{minipage}

\vspace{2\baselineskip} 
\begin{minipage}{\textwidth}
    \subsection{Downstream Agent Architecture Ablation}
    \begin{table}[H]\centering
        \caption{Downstream agent architecture ablation using Flan-T5 base as answering agent on the Full validation set.}\label{tab:m1_ablation}
        \begin{tabular}{lrrrrrrrrrr}\toprule
            &\multicolumn{4}{c}{F1} & &\multicolumn{4}{c}{EM} \\\cmidrule{2-5}\cmidrule{7-10}
            &Incomplete &Response &Complete &Recovery & &Incomplete &Response &Complete &Recovery \\\midrule
            Flan-T5-Small &41.4 &51.1 &53.6 &79.3 & &28.5 &35.3 &37.8 &73.0 \\
            Flan-T5-Base &53.0 &68.8 &71.7 &84.4 & &39.0 &54.3 &56.0 &89.7 \\
            Flan-T5-Large &59.8 &76.1 &81.8 &74.2 & &42.5 &58.0 &63.5 &73.8 \\
            Flan-T5-XL &62.3 &78.9 &82.9 &80.5 & &45.8 &60.8 &64.8 &78.9 \\
            Flan-T5-XXL &65.2 &78.9 &82.2 &80.6 & &50.5 &62.5 &65.8 &78.7 \\
            GPT-3.5 Turbo &60.3 &65.6 &67.7 &72.5 & &33.8 &36.8 &39.3 &54.5 \\
            GPT-4 &63.3 &68.7 &72.4 &59.1 & &39.3 &44.8 &46.8 &73.3 \\
            \bottomrule
        \end{tabular}
    \end{table}
\end{minipage}


\vspace{2\baselineskip} 
\begin{minipage}{\textwidth}
    \subsection{Alexpaca Training Hyperparameters}
    \begin{table}[H]\centering
        \begin{tabular}{ll}\toprule
            Examples & 500 \\
            Per Device Batch Size & 2 \\            
            Gradient Accumulation Steps & 16 \\
            Learning Rate & 2e-5 \\
            Weight Decay & 0 \\
            Warmup Ratio & 0.03 \\
            Learning Rate Schedule & Cosine \\
            Data Parallel & Full Shard Auto Wrap \\
            Random Seed & 0 \\
            \bottomrule
        \end{tabular}
    \end{table}
    We perform training on 2x NVIDIA A100 GPUs. We perform inference on 1x NVIDIA RTX A6000 with batch size 1. 
\end{minipage}
\end{document}